# Recognition Networks for Approximate Inference in BN2O Networks

**Quaid Morris**[*]
Gatsby Computational Neuroscience Unit
University College London
17 Queen Square, London, WC1N 3AR. England
quaid@gatsby.ucl.ac.uk

## Abstract

A recognition network is a multilayer perception (MLP) trained to predict posterior marginals given observed evidence in a particular Bayesian network. The input to the MLP is a vector of the states of the evidential nodes. The activity of an output unit is interpreted as a prediction of the posterior marginal of the corresponding variable. The MLP is trained using samples generated from the corresponding Bayesian network.

We evaluate a recognition network that was trained to do inference in a large Bayesian network, similar in structure and complexity to the Quick Medical Reference, Decision Theoretic (QMR-DT) network. Our network is a binary, two-layer, noisy-OR (BN2O) network containing over 4000 potentially observable nodes and over 600 unobservable, hidden nodes. In real medical diagnosis, most observables are unavailable, and there is a complex and unknown process that selects which ones are provided. We incorporate a very basic type of selection bias in our network: a known preference that available observables are positive rather than negative. Even this simple bias has a significant effect on the posterior.

We compare the performance of our recognition network to state-of-the-art approximate inference algorithms on a large set of test cases. In order to evaluate the effect of our simplistic model of the selection bias, we evaluate algorithms using a variety of incorrectly modelled selection biases. Recognition networks perform well using both correct and incorrect selection biases.

## 1 INTRODUCTION

We are interested in approximate inference in large discrete-valued Bayesian networks (BNs). Inference is the process of calculating posterior probability distributions over sets of unobserved variables given the states of the observed variables (ie the evidence). In large networks containing loops exact inference is often intractable. However, one can gain tractability at the expense of accuracy by doing approximate inference.

Two major classes of approximate inference methods are variational methods and Monte Carlo methods. Variational methods approximate the posterior using a member of a parameterised family of distributions. The approximating distribution is chosen by a deterministic algorithm that attempts to maximise a measure (or a bound thereof) of the goodness of fit to the true posterior. It is often difficult, however, to find goodness of fit measures and parameterised families which are both easily optimised and lead to accurate approximations. Furthermore, since the true posterior typically doesn't reside in the parameterised family, the accuracy of the approximation is limited. On the other hand, Monte Carlo methods can, in principle, approximate the posterior to arbitrary accuracy using samples from the Bayesian network. However, in practice, accurate approximations can require a prohibitively large number of samples.

In this paper, we propose recognition networks as an alternative method which has the potential to be both fast and accurate. Recognition networks, like variational methods, approximate the posterior with a member of a family of parameterised distributions. However the variational parameters are reoptimised for each new set of evidence whereas in recognition networks, the parameters are only optimised once. The recognition network parameters define a deterministic mapping from the evidence to the network's approximation of the posterior. If the deterministic mapping is sufficiently flexible, the posterior can be approxi-mated to arbitrary accuracy. Inference in a recognition network is done with a single feedforward pass.

---

[*] also affiliated with Department of Brain and Cognitive Sciences at MIT

We compare recognition networks to other approximate inference algorithms using a large binary, two layer, noisy-OR network (BN2O). This network was designed using published statistics to be similar to the Quick Medical Reference, Decision Theoretic (QMR-DT). The QMR-DT is a large Bayesian network that models medical diagnosis and has been widely used as a benchmark for approximate inference algorithms [Shwe and Cooper, 1991, D'Ambrosio, 1994, Murphy et al., 1999, Ng and Jordan, 2000]. The set of observable nodes in the QMR-DT are called findings[1] and the unobservable nodes are diseases.

When evaluating inference algorithms on the QMR-DT, one must be careful in one's choice of benchmarks. There is a difference between benchmarks which are externally defined, ie those that use realistic medical data, or internally defined, ie those that are defined by the QMR-DT without any reference to the medical domain. The difference arises partially because of invalid conditional independence assumptions encoded in the structure of the QMR-DT (see [Shwe et al., 1991] for a discussion of the effect of these simplifying assumptions). Externally defined benchmarks unfairly penalise [reward] particular inference algorithms for their failure [success] at overcoming these simplifications. On the other hand, internally defined benchmarks risk irrelevance to the medical domain.

The most commonly used benchmarks for the QMR-DT combine both internal and external elements. Though other benchmarks exist that are either fully internal [Frey et al., 2001] or external [Middleton et al., 1991], they have been less widely adopted. The internal/external benchmarks use sets of positive and negative findings from real patient cases (CPC cases) or pedagogical diagnostic problems manufactured by experts (SAM cases). However these benchmarks use the exact posterior marginals generated by Quickscore [Heckerman, 1989] as the gold standard rather than the reference diagnosis provided with the finding sets. Though the QMR-DT defined standard is more fair, one hopes that it isn't very different from what an externally defined gold standard would be.

There is, however, a potentially serious problem with using the Quickscore/CPC benchmark that, to our knowledge, has been previously overlooked. Generating the gold standard using the QMR-DT makes a strong and incorrect assumption about the process that selects the observed subset of finding nodes. During the diagnostic procedure, a series of decisions is made that determine what laboratory tests are ordered, what physical examinations are performed, and how the patient is interviewed. Clearly each decision is based, in part, upon the results of previously made investigations (ie the states of other finding nodes). However, Quickscore implicitly assumes that each of the decisions was made independently of the states of other findings. This difference is important because the diagnostic procedure introduces a selection bias. One aggregate effect of this bias is that a much larger proportion of the reported findings are positive than one should expect if there were no selection bias. We call this effect an *observation bias* toward positive findings. With this observation bias, unobserved finding nodes are more likely to be negative than they would be without the bias. This effect introduces an additional, and potentially large, source of variation between the QMR-DT gold standard and the reference diagnosis. Unfortunately the effect cannot be captured within the QMR-DT itself.

One can, however, augment the QMR-DT to model the observation bias. This augmentation adds an extra set of nodes that indicate whether or not the corresponding finding node was observed. The conditional probability tables associated with these new nodes define the level of observation bias assumed to be present. The posterior calculated over the disease nodes in the augmented network includes the effect of the assumed observation bias.[2] Unfortunately, due to the extra set of nodes, Quickscore can no longer be used to calculate exact posterior marginals. This requires us to introduce a new method of evaluating the approximate posteriors.

We evaluate approximate inference algorithms using purely internally defined benchmarks. Since neither the QMR-DT nor the associated CPC and SAM cases are publicly available, we generate our own BN2O with similar structure and statistics to the QMR-DT. We augment our BN2O network the same way we suggest doing for the QMR-DT. In our benchmark we use finding sets generated by forward sampling from our augmented network. Since we are generating our own finding sets, we can vary the level of observation bias present in the finding sets and measure the sensitivity of the performance of each approximate inference method to these variations. Ideally, an approximate inference method should be fairly insensitive to changes in the observation bias.

In the following section, we describe the QMR-DT

---

[1] A finding is something that a physician may discover about a patient and could include a patient's medical history, symptoms, physical signs or laboratory test results [Miller et al., 1982]

[2] Note that the actual level of observation bias present in a real medical case will not generally be available

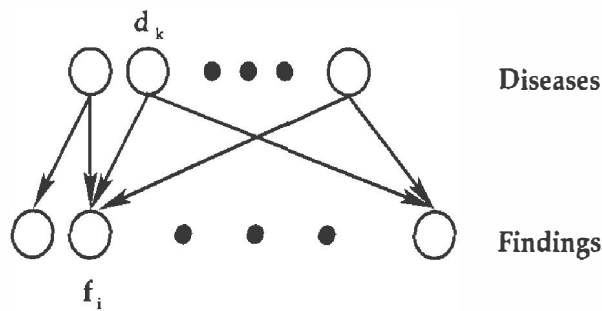

Figure 1: The graphical structure of the QMR-DT. There are approximately 600 (hidden) disease nodes and 4,000 (possibly visible) finding nodes.

and our BN2O network in greater detail. In section three, we describe our model of the observation process. In section four, we introduce recognition networks and describe how we implement them using multilayer perceptrons. In section five, we will describe a new method of comparing approximate inference algorithms on the QMR-DT. The final section of this paper is the experimental section in which we compare approximate inference algorithms using different observation biases.

## 2 QMR-DT

The QMR-DT network has a bipartite graph structure as shown in figure 1. Both the top-level disease nodes and the bottom-level finding nodes are binary valued. A positive finding almost always represents an abnormal state of the corresponding feature.

The probability distribution, $P(\mathbf{d}, \mathbf{f})$, represented by the QMR-DT factors as

$$P(\mathbf{d}, \mathbf{f}) = [\prod_{k=1}^{K} P(d_k)][\prod_{i=1}^{I} P(f_i \mid \pi_i)]$$

where $\pi_i$ are the parents of finding $i$. All parent nodes are diseases. The diseases are marginally independent and the findings are conditionally independent from one another given the states of the diseases. The conditional distribution of a finding is parameterised using the noisy-OR function. In this parameterisation, the conditional probability of a negative finding, $P(f_i = - \mid \mathbf{d})$, has the form

$$P(f_i = - \mid \mathbf{d}) = (1 - q_{i0}) \prod_{d_k \in \pi_i} (1 - q_{ik})^{d_k} \quad (1)$$

$$= e^{-\theta_{i0} - \sum_k \theta_{ik} d_k} \quad (2)$$

The parameters $q_{ik}$ are the probabilities that disease $k$ will cause finding $i$ by itself. The leak term $q_{i0}$ is the probability that a leak event will occur at finding $i$, ie the finding will be positive when none of its parents are on. Equation (2) is a useful reparameterisation of (1), using $\theta_{ij} = -\log(1 - q_{ij})$.

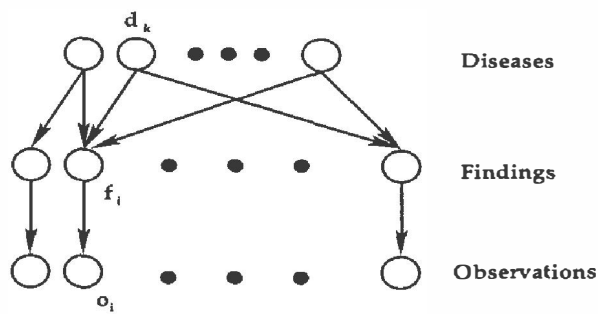

Figure 2: Augmented QMR-DT

We use published QMR-DT statistics to both build our BN2O network and to make some inferences about a real QMR-DT. From Shwe et al (1991) we know the following facts:

1. Disease configurations are sparse since $P(d_k = 1) \in [2 \times 10^{-5}, 2 \times 10^{-2}]$.

2. Leak events are rare since $q_{i0} \in [5.8 \times 10^{-8}, 0.153]$.

3. $q_{ik} \in \{0.025, 0.2, 0.5, 0.8, 0.985\}$.

4. Average number of findings connected to a disease node is 70.

Using fact 4 we can predict that each disease will cause roughly 35 positive findings.[3] Using this and facts 1 and 2, we can estimate that in fully observed samples from the QMR-DT, the number of positive findings will be a few hundred or less. This means that there are at least an order of magnitude more negative findings than positive ones.

On the other hand, in the CPC and SAM cases, the reported numbers of positive findings and negative findings are typically of the same order of magnitude (see [Middleton et al., 1991, Shwe and Cooper, 1991, Jaakkola and Jordan, 1999] for examples). This discrepancy suggests that the observation process reveals a much larger proportion of the positive findings than the negative findings. We will discuss the effect of this bias in the following section.

## 3 OBSERVATION BIAS

In realistic diagnostic problems, only a small subset of the finding nodes are observed. The selection of the observed nodes depends on a number of factors including patient's initial complaint, the internist's beliefs, and cost and morbidity of each test. This observation process is complex and difficult to model.

---

[3] If we assume that the average $q_{ik}$ for disease $k$ is 0.5

We model the observation process by adding an extra layer of observation nodes onto the QMR-DT, one for each finding node. The augmented network is shown in figure 2. Each observation node $o_i$ is the unique child of the corresponding finding node $f_i$. In any diagnostic problem, all of the observation nodes are assigned a value. If $F^+$ is the set of indices of the positive findings in a given case and $F^-$ contains the indices of the negative findings, then each observation node $o_i$ is assigned a value as follows:

$$o_i = \begin{cases} + & i \in F^+ \\ - & i \in F^- \\ ? & \text{otherwise} \end{cases}$$

The value ? (unknown) indicates that finding $i$ wasn't observed.

We use the same conditional probability table for each node $o_i$. We can fully specify this table using two parameters, $p^+$ and $p^-$, as follows:

$$\begin{aligned} P(o_i = ? \mid f_i = +) &= p^+ \\ P(o_i = ? \mid f_i = -) &= p^- \end{aligned} \quad (3)$$

Roughly speaking, for a particular case, $p^+$ [$p^-$] is the proportion of all the possible positive [negative] findings that remain unobserved.

Under this observation process, the fact that a finding is unobserved can be informative about its hidden state. Using equation (3) and the structure of the augmented QMR-DT, we can write the probability that finding $i$ is unobserved given disease vector $\mathbf{d}$ as

$$\begin{aligned} P(o_i = ? \mid \mathbf{d}) = \\ p^+ P(f_i = + \mid \mathbf{d}) + p^- P(f_i = - \mid \mathbf{d}) \end{aligned} \quad (4)$$

Using equation (4), we can write the posterior distribution over $\mathbf{d}$ given $o_i = ?$ as

$$\begin{aligned} P(\mathbf{d} \mid o_i = ?) \propto \\ P(\mathbf{d}) \left[ p^+/p^- P(f_i = + \mid \mathbf{d}) + P(f_i = - \mid \mathbf{d}) \right] \end{aligned} \quad (5)$$

Equation (5) demonstrates that the observation process influences the posterior only through the ratio $p^+/p^-$. We call this ratio the *observation bias ratio*. If there are any unobserved findings then the posterior for the unaugmented QMR-DT will be different than the posterior for the augmented QMR-DT.[4]

---

[4] except in the unlikely case where $p^+/p^- = 1$

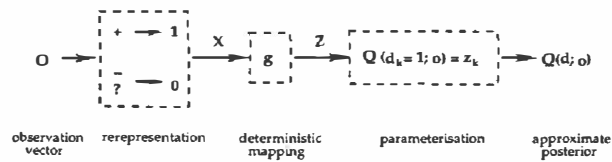

Figure 3: Process by which a recognition network generates an approximate posterior from an observation vector.

The observation bias ratio has a nice interpretation. Since in realistic cases almost all possible negative findings are hidden (ie $p^- \approx 1$), therefore $p^+/p^- \approx p^+$. The observation bias ratio is roughly equal the proportion of the possible positive findings that remain hidden.

As this interpretation makes clear, the observation bias ratio will change from problem to problem and even during the process of the diagnostic procedure. As such, approximate inference algorithms which can make use of information about the observation process should be robust to slightly inaccurate information. Our experiments test the sensitivity of inference algorithms to changes in the assumed values of $p^+$ and $p^-$.

Recognition networks incorporate the observation process in a natural way. The network parameters are optimised using samples from the augmented version of the BN2O network. These samples include the effect of the observation process. Therefore, the network is trained to do inference assuming a particular observation bias ratio. In the following section, we describe recognition networks in greater detail.

## 4 RECOGNITION NETWORKS

A recognition network is a deterministic function that maps an observation vector $\mathbf{o}$ to an approximate posterior distribution over the disease nodes $\mathbf{d}$. Figure 3 shows the whole process. First, the observation vector $\mathbf{o}$ is rerepresented as a binary vector $\mathbf{x}$. The input vector has elements $x_i = 1$ only if $o_i = +$ otherwise $x_i = 0$. In other words, our current implementation of recognition networks doesn't distinguish between negative and unobserved findings. A parameter vector $\mathbf{z}$ is then generated by evaluating the parameterised, vector-valued function $g(\mathbf{x}; \Omega)$. The tunable parameters, $\Omega$, of this function determine the behaviour of the recognition network. Each element $z_k$ of the parameter vector is interpreted as the marginal posterior probability that disease $k$ is present, ie

$$Q(d_k = 1; \mathbf{o}) = z_k$$

The distribution over all the diseases $Q(\mathbf{d}; \mathbf{o})$ can be written as

$$Q(\mathbf{d}; \mathbf{o}) = \prod_k Q(d_k; \mathbf{o}) = \prod_k z_k^{d_k}(1-z_k)^{(1-d_k)}$$

We use two styles of parameterisations of the vector-valued functions: logistic regression (LR) networks and multilayer perceptrons (MLP). The following equations specify the activities of the units in the hidden ($\mathbf{y}$) and output ($\mathbf{z}$) layers of the MLP given the vector of input activities ($\mathbf{x}$):

$$y_j = \tanh(b_j + \sum_i v_{ji} x_i) \quad (6)$$

$$z_k = \sigma(a_k + \sum_i w_{ki} x_i + \sum_j u_{kj} y_j) \quad (7)$$

where $\sigma(z) = (1 + \exp(-z))^{-1}$ is the logistic function and $\mathbf{a}$ and $\mathbf{b}$ are vectors of bias terms. The parameter set $\Omega = \{U, V, W, \mathbf{a}, \mathbf{b}\}$ contains all three weight matrices and the two vectors of biases.

The LR networks are MLPs without a hidden layer, ie the activity of the $k$-th output unit, $z_k$, is

$$z_k = \sigma(a_k + \sum_i w_{ki} x_i)$$

We use samples, $(\mathbf{d}^{(n)}, \mathbf{o}^{(n)})$, from our augmented BN2O network to train the recognition network. The reference diagnosis, $\mathbf{d}^{(n)}$, is used as the target vector. The error function we minimise, $E(\Omega)$, where $\mathbf{z}^{(n)}$ is the parameter vector generated using $\mathbf{o}^{(n)}$, is:

$$E(\Omega) = -\sum_n \sum_k d_k^{(n)} \log z_k^{(n)} + (1-d_k^{(n)})\log(1-z_k^{(n)}) \quad (8)$$

$E(\Omega)$ is the cross-entropy between the set of reference diagnoses and the approximate posterior $Q(\mathbf{d}; \mathbf{o}^{(n)})$. In the large sample limit, equation (8) is minimised when

$$z_k^{(n)} = P(d_k = 1 \mid \mathbf{o}^{(n)})$$

Since MLPs are universal approximators [Lapedes and Farber, 1988], it is possible, in principle, given enough hidden units and incorporating negative evidence to train the MLP to predict the exact posterior marginals.

## 5 EVALUATION

Inference methods for the QMR-DT are typically evaluated by comparing their approximate posteriors against the exact posterior marginals calculated using the Quickscore algorithm [Heckerman, 1989]. However, since Quickscore ignores unobserved findings, it generates incorrect posterior marginals when the observation bias ratio isn't one. This section describes how to compare approximate posteriors using lists of diagnoses.

### 5.1 D-LISTS

A diagnosis list ($D$-$list$) is an ordered set of disease configurations (aka diagnoses) $\{\mathbf{d}^1, \mathbf{d}^2, \ldots, \mathbf{d}^M\}$. The order of the D-list is important since in a clinical setting only diagnoses near the top of the list will ever be considered.

In this paper, the D-list assigned to an approximate posterior is the $N$ most probable diagnoses under that posterior. Since we only use fully factorised posteriors, we can use best-first search to generate this list.

### 5.2 EVALUATING D-LISTS

We score a diagnosis, $\mathbf{d}$, using its joint probability, $P(\mathbf{d}^m, \mathbf{o}^{(n)})$, with the observation vector $\mathbf{o}^{(n)}$. Since this score is equal to the posterior times a factor that doesn't depend upon $\mathbf{d}$, ie

$$P(\mathbf{d}, \mathbf{o}^{(n)}) = P(\mathbf{d} \mid \mathbf{o}^{(n)}) P(\mathbf{o}^{(n)})$$

we can use it to rank all diagnoses generated for the same evidence vector. The score of the reference diagnosis, $P(\mathbf{d}^{(n)}, \mathbf{o}^{(n)})$ provides a good benchmark to compare other diagnoses against.

We use the *average cumulative ratio curve* to measure the expected quality of a D-list over a distribution of observation vectors. The cumulative ratio curve of a D-list for $\mathbf{o}^{(n)}$ is defined for ranks 1 up to the length of the list. The value $C(r, \mathbf{o}^{(n)})$ of the cumulative ratio curve at rank $r$ is

$$C(r, \mathbf{o}^{(n)}) = Z(\mathbf{o}^{(n)})^{-1} \sum_{m=1}^r P(\mathbf{d}^m, \mathbf{o}^{(n)})$$

where $Z(\mathbf{o}^{(n)})$ is a normalising factor. This factor ensures that the curves for different observation vectors all have the same scale. We set $Z(\mathbf{o}^{(n)})$ to be

$$Z(\mathbf{o}^{(n)}) = \max\left[P(\mathbf{d}^*, \mathbf{o}^{(n)}), P(\mathbf{d}^{(n)}, \mathbf{o}^{(n)})\right]$$

where $\mathbf{d}^*$ is the diagnosis with the highest joint probability with $\mathbf{o}^{(n)}$ among all of the D-lists generated by the different algorithms being compared. Note that this ensures

$$C(1, \mathbf{o}^{(n)}) \leq 1$$

A good cumulative ratio curve starts near 1 and grows quickly. The average cumulative ratio curve is the average across the observation vectors within a test set of the individual cumulative ratio curves.

# 6 EXPERIMENTS

## 6.1 METHODS

We compare recognition networks against one variational method and one Monte Carlo method. We only run the competing methods on the unaugmented BN2O network. The variational method, JJ99 [Jaakkola and Jordan, 1999] has no obvious extension to the augmented network. The Monte Carlo method, AIS-BN [Cheng and Druzdzel, 2000], could be run on an augmented QMR-DT, how doing so would drastically increase its time complexity.

### 6.1.1 JJ99

Jaakkola and Jordan (1999) proposed the following method for approximate inference in the QMR-DT. They approximate the log posterior marginal odds of disease $k$,

$$l_k = \log \frac{P(d_k = 1 \mid \mathbf{o})}{P(d_k = 0 \mid \mathbf{o})}$$

with

$$\hat{l}_k = p_k - \theta_{i0} - \sum_{j \in F^-} \theta_{jk} + \sum_{i \in F^+} \theta_{ik}\xi_i$$

where $p_k = \log[P(d_k = 1)/P(d_k = 0)]$ is the log prior odds and the $\theta$ parameters are defined as in equation (2). The $\xi_i$ values are strictly positive variational parameters chosen to globally minimise a convex upper bound on the probability of $\mathbf{o}$. We use a standard nonlinear optimisation routine to do the minimisation.

### 6.1.2 AIS-BN

AIS-BN is an importance sampling method with an adaptive proposal distribution. Following Cheng and Druzdzel's (2000) empirical tests, we use a two phase version of AIS-BN. In the initial phase the proposal distribution adapts and in the second phase the proposal distribution is fixed and samples from it are used in the estimation of the posterior marginals. We used the parameter settings suggested by Cheng and Druzdzel, however we only use one of the two suggested initialisation heuristics for the proposal distribution. The initial proposal distribution, $\tilde{P}$, before any adaptation, is set as follows:

$$\tilde{P}(d_k = 1) = \begin{cases} P(d_k = 1) & \text{if } P(d_k = 1) > 0.04 \\ 0.04 & \text{otherwise} \end{cases}$$

Cheng and Druzdzel also suggest setting $\tilde{P}(d_k = 1) = 0.5$ in some cases, however initial tests showed that using this heuristic results in significantly worse performance. For each test case, we use 25,000 samples in the initial training phase and 75,000 samples in the second phase. The same samples are used to estimate both $P(d_k, \mathbf{o})$ and $P(\mathbf{o})$.

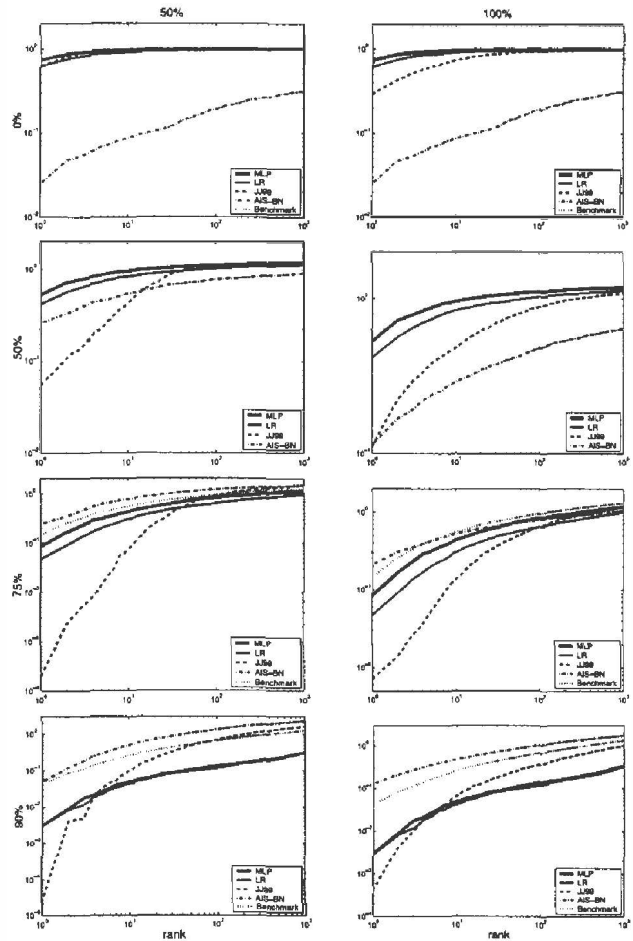

Figure 4: Average cumulative ratio curves for the eight test sets. There is one column for each value of $p^-$ and one row for each $p^+$. Notice the different scales.

### 6.1.3 MLP

We train our recognition networks using stochastic gradient descent to minimise equation (8). Our training algorithm incorporates an adaptive global learning rate $\eta^{(t)}$, a momentum term $\mu = 0.95$, and error centering [Schraudolph, 1998]. The update rule for an arbitrary, non-bias weight $\omega$ from $\Omega$ is

$$\Delta\omega^{(t+1)} = \eta^{(t)}\frac{dE}{d\omega} + \mu\Delta\omega^{(t)} - \langle\Delta\omega\rangle^{(t)}$$

where $\langle\Delta\omega\rangle^{(t)}$ is an estimate at the $t$-th mini-batch of the mean value of $\Delta\omega$. For bias weights, the update rule is the same except the $\langle\Delta\omega\rangle^{(t)}$ term isn't subtracted off.

We trained both an LR network and an MLP on samples from our augmented BN2O network using $p^+ = 0.5$ and $p^- = 1.0$. We first trained the LR network first using $10^7$ samples. The weight matrix, $W$, of the LR network became the fixed input-output weights for the MLP. Our MLP contains 1000 hidden units, The weights to the hidden units were trained using $10^7$ samples.

## 6.2 RESULTS

### 6.2.1 Benchmark Diagnostic Problems

In this section, we describe how we generate the benchmarks we use to compare the inference algorithms. We generate eight benchmark test sets by forward sampling from our augmented BN2O network with different values of $p^+$ and $p^-$. Each benchmark test set contains exactly 1000 test cases.

We attempt to generate benchmarks of similar difficult as the CPC cases. To this end we only generate reference diagnoses containing exactly five diseases by sampling from $P(\mathbf{d} \mid \sum_k d_k = 5)$, ie the disease prior conditioned on there being exactly five diseases in the reference diagnosis. We generate the observation vector $\mathbf{o}^{(n)}$ by clamping the disease nodes in configuration $\mathbf{d}^{(n)}$ and forward sampling from our augmented BN2O network.

We generate the eight different benchmarks by using all combinations of four different values of $p^+$ and two different values of $p^-$. The value of $p^+$ is chosen from $\{0, 0.5, 0.75, 0.9\}$. There are 100 positive findings on average in the test cases for which all positive findings are observed.

The value of $p^-$ is either 0.5 or 1. The $p^- = 1$ benchmark is designed to model realistic diagnostic problems. However, these benchmarks give recognition networks an unfair advantage since neither JJ99 nor AIS-BN incorporate information about the observation process. As we have no other convenient method to provide JJ99 and AIS-BN with this information, we use $p^- = 0.5$ benchmarks to ensure that there is at least one benchmark where the observation process doesn't effect the posterior (ie $p^+/p^- = 1$). However, the $p^- = 0.5$ benchmarks are extremely unrealistic since each test case contains over 2000 negative findings.

### 6.2.2 Results

Figure 4 shows the average cumulative ratio curves for test set. Bad performance on some test sets could either be due to an incorrect observation bias ratio or because the inference problem is particularly difficult. To distinguish between these two explanations, we include benchmark curves when $p^+ \neq 0.5$. These curves were generated by other logistic regression networks trained on samples from the augmented BN2O network with the same value of $p^+$ as the test cases.

Both the LR and MLP networks perform very well on all but one of the test sets. The MLP performs the best of all the methods when the assumed observation bias equals the actual observation bias and is competitive with the benchmark curve on the $p^+ = 0$ and $p^+ = .75$ cases. Surprisingly, the LR network performs almost as well as the MLP in all cases despite using only one third as many parameters as the MLP. Most importantly, both recognition networks have good cumulative ratio curves for low ranks, ie they place good diagnoses high up in their D-lists. AIS-BN seems particularly suited for approximate inference when only a few positive findings are available.

## 7 DISCUSSION

In summary, we have made two main contributions which include introducing MLPs for approximate inference. We are however, not the first to use recognition networks for approximate inference.

Recognition networks are a type of *recognition model*. Our logistic regression network is exactly the same as the single layer recognition model in the Helmholtz machine [Dayan et al., 1995, Hinton et al., 1995] and is trained using a similar algorithm. Our contribution to this earlier work is the addition of a hidden layer to the recognition model.

Our second contribution lies in identifying the importance of the observation process. Observation processes are important in any inference problem where some of the potentially observable variables are hidden. We've showed that ignoring or incorrectly modelling even a very basic type of selection bias can affect the accuracy of approximate inference. In particular, we have shown a non-trivial and algorithm-specific effect of observation bias.

One obvious extension of this work would be to incorporate negative evidence into the recognition network. This would require using 1-of-$N$ coding for the input units. Recognition networks could be used for inference in arbitrary Bayesian networks. However, this may require 1-of-$N$ coding for both the input and output units.

## ACKNOWLEDGMENTS

I would like to thank Peter Dayan and Geoffrey Hinton for their significant contributions. Additionally I would like to thank Brendan Frey, Kevin Murphy, Relu Patrascu, Sam Roweis, and especially Yee Whye

Teh for helpful discussions. This work was generously funded by the Gatsby Charitable Foundation.